\documentclass{article}

\usepackage{PRIMEarxiv}

\usepackage[utf8]{inputenc} % allow utf-8 input
\usepackage[T1]{fontenc}    % use 8-bit T1 fonts
\usepackage{hyperref}       % hyperlinks
\usepackage{url}            % simple URL typesetting
\usepackage{booktabs}       % professional-quality tables
\usepackage{amsfonts}       % blackboard math symbols
\usepackage{nicefrac}       % compact symbols for 1/2, etc.
\usepackage{microtype}      % microtypography
\usepackage{lipsum}
\usepackage{fancyhdr}       % header
\usepackage{graphicx}       % graphics
\graphicspath{{media/}}     % organize your images and other figures under media/ folder
\usepackage{times}
\usepackage{latexsym}
\usepackage{amsfonts}
\usepackage{graphicx}
\usepackage{caption}
\usepackage{subcaption}
\usepackage{arydshln}
\usepackage{natbib}
\usepackage[utf8]{inputenc}
\usepackage[english]{babel}
\title{Multi-hop Question Answering over Knowledge Graphs using Large Language Models
}

\author{
  Abir Chakraborty \\
  Microsoft \\
  \texttt{Abir.Chakraborty@microsoft} \\
}

\begin{document}
\maketitle

\begin{abstract}
Knowledge graphs (KGs) are large datasets with specific structures representing large knowledge bases (KB) where each node represents a key entity and relations amongst them are typed edges. Natural language queries formed to extract information from a KB entail starting from specific nodes and reasoning over multiple edges of the corresponding KG to arrive at the correct set of answer nodes. Traditional approaches of question answering on KG are based on (a) semantic parsing (SP), where a logical form (e.g., S-expression, SPARQL query, etc.) is generated using node and edge embeddings and then reasoning over these representations or tuning language models to generate the final answer directly, or (b) information-retrieval based that works by extracting entities and relations sequentially. In this work, we evaluate the capability of (LLMs) to answer questions over KG that involve multiple hops. We show that depending upon the size and nature of the KG we need different approaches to extract and feed the relevant information to an LLM since every LLM comes with a fixed context window. We evaluate our approach on six KGs with and without the availability of example-specific sub-graphs and show that both the IR and SP-based methods can be adopted by LLMs resulting in an extremely competitive performance.
\end{abstract}

% keywords can be removed

\section{Introduction}
Predicting factual answers from a knowledge base (KB) is a key capability that finds applications across all industries and different machine learning (ML) approaches exist to tackle different types of knowledge base. In most of the cases, the knowledge base is a combination of structured or unstructured database and the structured database further can be either relational or graph-based \citep{miller-etal-2016-key, saxena-etal-2020-improving, yu-etal-2018-spider, yu-etal-2019-sparc, gu2023interleaving}. For unstructured data it can be large documents \citep{rajpurkar-etal-2016-squad, kwiatkowski-etal-2019-natural} (i.e., machine reading comprehension task), open domains \citep{DBLP:journals/corr/abs-2101-00774} where we need to identify the relevant documents first before generating the final answer and finally closed-book methods \citep{roberts-etal-2020-much} that exploit the knowledge stored in the model parameters as part of training.

Graph based knowledge-base (also called knowledge-graph) can originate either because of the nature of the data (e.g., Facebook social graph where users and activities are the nodes of a large graph) or by processing large knowledge base where relations amongst KB entities (person, event, etc.) are explicitly captured for better discovery. Some of the well-known large-scale KGs are Wikidata \citep{wikidata_2014} with more than 733M triples and 58M nodes, Freebase \citep{freebase_2008} (now deprecated) and DBpedia \citep{dbpedia_2015} that has more than 220 M entities and 1.45 billion triples\footnote{https://www.dbpedia.org/resources/knowledge-graphs/}. Answering questions posed over these large KGs requires successful resolution of many sub-tasks, namely, (1) mention detection, which is similar to a nested Named Entity Recognition (NER) task to identify spans (possible discontinuous) of the question that contains the key entity information, (2) entity recognition, i.e., mapping these spans to the nodes of the KG, (3) sub-graph extraction, i.e., extracting multi-hop neighborhood around these key entities for reducing the complexity of searching the answer node, and finally, (4) identify the right path in this sub-graph. As an example, for the question, “what napa county wine is 13.9 percent alcohol by volume?” first we need to identify the right span, "napa county" and then mapping to related entities, e.g., napa valley or napa airport.

As mentioned earlier, there are two distinct approaches to question answering on a KG. The first one is based on Semantic Parsing (SP), where a logical form (LF) is generated first (say, in SPARQL or S-expression) and the generated LF is subsequently executed on the graph database to get the final result. Till the time the LF is executed there is no way of knowing whether the LF is even syntactically accurate. The second approach is in the category of Information Retrieval (IR) and it tries to directly generate the answer without creating an LF first, and in general, this approach is lagging behind the first one in terms of accuracy. The IR based approach proceeds by extracting relevant entities from the graph and ranking them to get the top answer. All sequence-to-sequence approaches can be considered part of IR based approach as it does not generate a query but directly tries to generate the answer. The advantage of SP based approach is that there is no need to extract sub-graphs for each example, which can be challenging as we never know the required number of hops a priori. There are several SP datasets where sub-graphs are not available, e.g., LC-QuAD \citep{trivedi2017lc}, LC-QuAD 2.0 \citep{lcquad_2}, GrailQA \citep{grailqa_21} and KQAPro \citep{cao-etal-2022-kqa} and the only way of arriving at the answer node is to map a natural language question to the corresponding LF (SPARQL or S-expression). 

As the large language models (LLMs) are changing the landscape of NLP domain setting new milestones on the traditional tasks \citep{han2023information, zhao2023survey} it is worth evaluating them on more challenging tasks like question answering on KB (relational or graph database). These models mostly work in a zero-shot or few-shot setup \citep{ouyang2022training} where typically an instruction prompt is designed without any model fine-tuning. However, LLMs can be prone to generating unsubstantiated information \citep{10.1609/aaai.v37i11.26535} unless well-defined guardrails are provided in the prompt and even then it may lack domain-specific or latest knowledge \citep{schick2023toolformer, peng2023check}. Thus, we need to augment the prompt with the relevant part of the KB so that questions can be answered by utilizing the content of the KB and not relying on what the LLM has learnt during its training.

This brings us to the question of what is the best representation of the (partial) external KB that can be included in the prompt. This will be decided based on the approach we take. Thus, for the IR based approach we need to pass either a list of entities (nodes) or relations (edges) while identifying a specific node or edge and a linearized version of the sub-graph (verbalized triples) while extracting the right path. Similarly, for the SP based approach (since we do not have access to the graph data) we can only pass a linearized schema of the graph database, i.e., description of the Ontology classes, their relations and attributes, which in turn is another graph (albeit much smaller than the original graph). Another consideration comes from the fact that every LLM has an associated context window length (GPT-3.5 originally has a limit of 4096 tokens) that has to accommodate instructions and examples along with the part of the KB. Thus, depending upon the size of the sub-graph we may have to resort to Retrieval Augmented Generation (RAG) \citep{gao2024retrievalaugmented} that will dynamically extract the relevant part of the sub-graph. Hereafter, we call the first approach IR-LLM where graph snippets are part of the context and the second one SP-LLM where only the graph schema is provided. Notably, the SP-LLM approach would not require RAG if we have a concise description of the graph schema.

In this work we show how LLMs (we use only {\tt GPT-3.5-turbo} for all our experiments) can be utilized to answer KG based questions requiring multiple hops. However, contrary to a unified approach as espoused by previous studies \citep{jiang-etal-2023-structgpt} we show that we need to adopt different strategies for different KG dataset. We evaluate our approach on three KGs where the sub-graphs associated with each example is available and hence we adopt the IR-LLM strategy and on the other three KGs where no such sub-graph is available we apply the SP-LLM strategy. The organization of the rest of the paper is as follows. In the next section we provide a detailed literature survey on the ML-based techniques employed for KG question-answering followed by the more recent LLM based approaches. Next, we present our approach for IR-LLM and SP-LLM. Subsequently, our LLM based predictions and comparisons with other baseline methods are discussed. Finally, conclusions are drawn and scope for future works is outlined.

\begin{table*}
\centering
\begin{tabular}{l|cccc|c}
\hline
Datasets & \#Train & \#Valid & \#Test & Max \#hop & Type \\
\hline
WebQSP,~\cite{Berant2013SemanticPO} & 2,848 & 250 & 1,639 & 2 & IR \\
MetaQA-3hop,~\citep{zhang2017variational} & 114,196 & 14,274 & 14,274 & 3 & IR \\
CWQ,~\cite{talmor-berant-2018-web} & 27,639 & 3,519 & 3,531 & 4 & IR \\ \hdashline
LC-QuAD-v1.0,~\cite{trivedi2017lc} & 4,000 & - & 1,000 & 2 & SP\\
LC-QuAD-v2.0,~\cite{lcquad_2} & 24,180 & - & 6,046 & 3 & SP \\
KQAPro,~\cite{cao-etal-2022-kqa} & 94,376 & 11,797 & 11,797 & $\ge 5$ & SP \\
\hline
\end{tabular}
\caption{Statistics of the datasets considered in this study in terms of the number of examples in the train, validation and test split, maximum number of hops required to answer a question and finally, the type of approach required to generate an answer.}
\label{tab:stat}
\end{table*}

% \begin{figure}
%   \centering
%   \fbox{\rule[-.5cm]{4cm}{4cm} \rule[-.5cm]{4cm}{0cm}}
%   \caption{Sample figure caption.}
%   \label{fig:fig1}
% \end{figure}

% \subsection{Tables}
% \lipsum[12]
% See awesome Table~\ref{tab:table}.

% \begin{table}
%  \caption{Sample table title}
%   \centering
%   \begin{tabular}{lll}
%     \toprule
%     \multicolumn{2}{c}{Part}                   \\
%     \cmidrule(r){1-2}
%     Name     & Description     & Size ($\mu$m) \\
%     \midrule
%     Dendrite & Input terminal  & $\sim$100     \\
%     Axon     & Output terminal & $\sim$10      \\
%     Soma     & Cell body       & up to $10^6$  \\
%     \bottomrule
%   \end{tabular}
%   \label{tab:table}
% \end{table}

\section{Related Work}
As mentioned earlier, there are two broad categories of work for KG question-answering (KGQA), namely, IR and SP based approaches. In the SP based approaches, there are different strategies for generating the LF. \citet{chen-etal-2021-retrack} uses multiple components, namely, retriever, transducer and checker, to retrieve relevant KB items and generating LF with syntax check. \citet{ye-etal-2022-rng} takes the approach of first ranking candidate LFs and then generates the final LF using these ranked ones and the original question. Another iterative process is introduced by \cite{gu2022arcaneqa} where intermediate LFs are executed and based on their results the final LF is created. A multi-grained retrieval approach is proposed by \citet{shu-etal-2022-tiara} that focuses on the most relevant entities, LFs and schema items. It is to be noted that till the time the LF is not executed there is no feedback on its accuracy and one needs an external executor such as a SPARQL server to execute the predicted LFs.

IR based methods, on the other hand, directly generate the final answer without relying on any LF generation or having no dependency on an external LF executor. \citet{sun2019pullnet} first extracts a question-specific sub-graph and then uses graph convolution network to decide the importance of each node in the sub-graph and to predict the final answer entities. \citet{saxena2022sequencetosequence} takes the sequence-to-sequence approach of directly generating the answers based on the T5 model \citep{raffel2023exploring} while introducing additional tasks of head and tail prediction. \citet{oguz-etal-2022-unik} also uses the sequence-to-sequence framework for open-domain question answering but like \cite{ye-etal-2022-rng} they first retrieve candidate triples from a KB and then generate the final answer using these triples and the input question. While they do not need external LF evaluator, IR based methods generally fall behind SP based methods on public benchmarks \citep{talmor-berant-2018-web, grailqa_21, Gu2022KnowledgeBQ}.

\section{Approach}
A knowledge graph is typically represented as a set of triples where each triple consists of a head entity, a relation and a tail entity, i.e., $\mathcal{G} = \left\{ (e, r, e') | e, e' \in \mathcal{E}, r \in \mathcal{R} \right\}$ where $\mathcal{E}$ and $\mathcal{R}$ are the set of entities and relations, respectively. In this work, all the edges are considered bi-directional, i.e., a 1-hop neighborhood of entity $e$ is defined as $\{e' | (e', r, e) \in \mathcal{G} \cup (e, r, e') \in \mathcal{G}\}$. If we denote a subset of $\mathcal{G}$, by $\mathcal{G}_s$ then the corresponding edges and vertices are denoted by $\mathcal{R}_s \in \mathcal{R}$ and $\mathcal{E}_s \in \mathcal{E}$, respectively. 

\begin{figure*}
     \centering
     \begin{subfigure}[b]{0.4\textwidth}
         \centering
         \includegraphics[width=\textwidth]{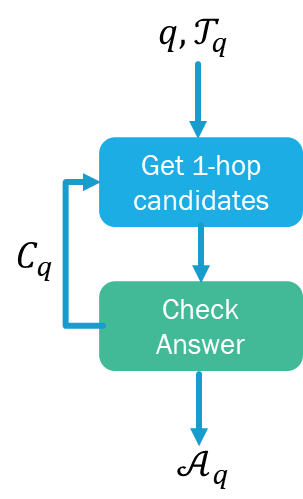}
         \caption{The main flow that invokes 'get 1-hop candidates' component.}
         \label{fig:finea}
     \end{subfigure}
     \hfill
     \begin{subfigure}[b]{0.4\textwidth}
         \centering
         \includegraphics[width=\textwidth]{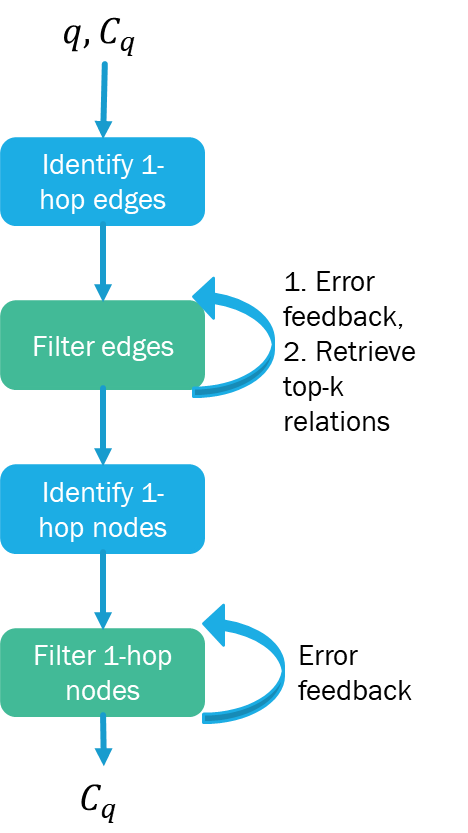}
         \caption{Sub-components that get 1-hop nodes.}
         \label{fig:fineb}
     \end{subfigure}
     \caption{Different components for question-answering from a KB. Left subfigure: the main flow that invokes 'get 1-hop candidates' component.   }
        \label{fig:fine}
\end{figure*}

\begin{figure*}
 \centering
 \includegraphics[width=0.5\textwidth]{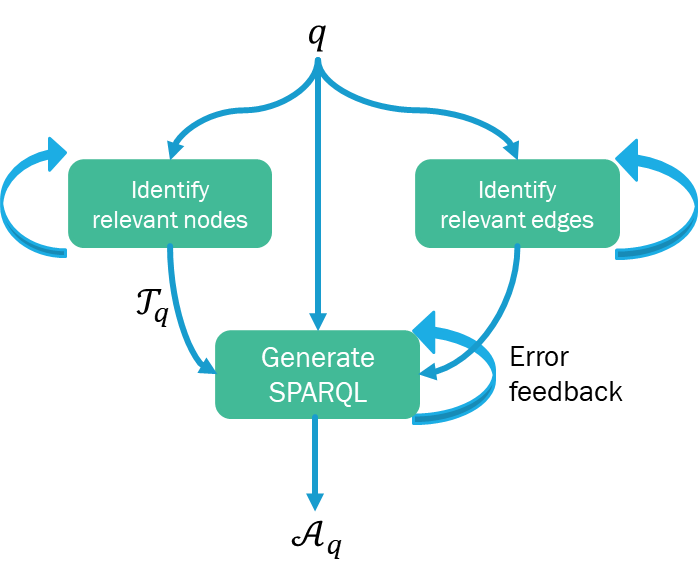}
 \caption{The flow for SP-LLM that generates a SPARQL query from the given question. There are three skills, (1) entity identification, (2) predicate identification and finally, (3) SPARQL generation.}
 \label{fig:sp}
\end{figure*}

Given a natural language question $q$ and a KG $\mathcal{G}$, the objective of KGQA is to find answer nodes (entities) in the KG, denoted by the answer set, $\mathcal{A}_q$. Following previous work \citep{sun-etal-2018-open, sun2019pullnet}, for the IR-LLM approach, it is assumed that the key entities mentioned in the question are known (called {\it topic entities}, denoted by $\mathcal{T}_q \in \mathcal{E}$). Thus, answering the question amounts to finding entity nodes that are within k-hop neighborhood of $\mathcal{T}_q$. Previous approaches have followed a two-step procedure, namely, retrieval and reasoning \citep{jiang2023unikgqa, jiang-etal-2023-structgpt}. The retrieval step extracts a subgraph $\mathcal{G}_q \in \mathcal{G}$ so that $\mathcal{A}_q \in \mathcal{E}_q$. Subsequently, a reasoning model filters $\mathcal{E}_q$ to extract $\mathcal{A}_q$. \citet{jiang2023unikgqa} further aggregates nodes with common ancestor to a single abstract node for simplifying subsequent computations.

\subsection{IR-LLM and SP-LLM}
In case of IR-LLM, our approach is broadly similar to the above as we also retrieve a part of the graph and ask LLM to reason over it but we do it in multiple steps. Since different LLMs have different permissible context window length (4096 for GPT-3.5-turbo to 8k, 16k and 32k for different GPT-4 variants) we cannot retrieve the entire subgraph and pass all the information to the LLM for reasoning. Instead, we proceed as shown below:
\begin{enumerate}
\item Identify all the 1-hop relations from the subgraph for a given topic entity set, $\mathcal{T}_q$.
\item Pass all the retrieved relations to the LLM for filtering and return top-k filtered relations (typically k = 1). We call it relation extraction skill-1 for easy reference.
\item If the retrieved relations are valid (present in the graph) then identify all the nodes associated with these relations. At this level even 2-hop nodes are retrieved if it is found that some of the 1-hop nodes are Compound Value Type (CVT) as it happens for the Freebase graph. 
\item If all the retrieved nodes are valid (present in the graph) then pass all the retrieved nodes to the LLM and decide whether any of these candidate nodes, $\mathcal{C}_q$, can be one of the answer nodes, $\mathcal{A}_q$. We call it entity filtering skill.
\end{enumerate}
These steps 1-4 are shown in Fig.~\ref{fig:fineb}. If the LLM decides that there is no suitable answer at step-4 then steps 1-3 are repeated with the current set of candidate nodes, $\mathcal{C}_q$ (as shown in Fig.~\ref{fig:finea}. Thus, in the above flow, step (2) and (4) are LLM based and step (1) and (3) are graph element retrieval that does not require any LLM call. 

For the SP-LLM approach, we introduce three skills (Fig.~\ref{fig:sp}), to (a) identify the relevant nodes from the description of all nodes in a large graph, (2) identify the relevant edges/predicates from their descriptions and (3) finally to construct the SPARQL query using the question, retrieved nodes and edges.

Since there are known issues (e.g., grounded response generation) with LLM generated responses, we also apply several checks for some of these skills. The first one is a feedback mechanism where if an identified relation (edge) or entity (node) is not present in the graph then an error message is sent back to the LLM as additional context in the prompt. The default flow does not have this context and it activates only after the first error and persists for user-specified number of retries. In addition, each LLM call is enriched with Few-shot examples where again the default behavior is zero-shot but additional examples are provided if LLM generated outputs are not present in the graph. In addition, if the first relation extraction skill hallucinates (generates relations that are not present in the graph) then there is another relation extraction skill that generates multiple candidate relations instead of just one. The rest of the flow remains same as before.  

All the three skills mentioned above can work as a self-contained prompt (where the number of candidate relations/entities is small enough to fit within the context length) or as a Retrieval Augmented Generation (RAG) where the candidate relations/entities are chunked and converted into vectors using {\tt text-embedding-ada-002} and retrieved based on their similarity with a composite prompt that captures both the question $q$ and topic entities, $\mathcal{T}_q$ (default) and few-shot examples (when they are utilized). The decision to use RAG depends upon the example and the number of connected nodes/edges and is created on-the-fly as required. 

\section{Datasets}
In this section we discuss the datasets used in this study in details including the nuances of each one and why it is difficult to have a consistent approach across all of them. 

\subsection{WebQSP}
WebQSP~\citep{Berant2013SemanticPO, Yih2016TheVO} is an English language question-answering dataset based on Freebase KG requiring at most two hops. The number of train, validation and test examples are shown in Table~\ref{tab:stat}. Here we have a single graph with 106 million triples instead of individual sub-graphs for each example. There are 1,441,420 entities and 6,102 relations in the corresponding graph making it difficult to apply RAG on the entity set. The entities are coded by integer ids from their Machine Identifiers or MIDs, e.g., id 15 has MID m.06w2sn5, which represents Justin Bieber (written as ‘Justin beiber’ though). Similarly, relations have their MIDs, e.g., m.01l6dm is location.country.currency\_used. An entity can be a simple or Compound Value Type (CVT) and the CVT entities, by design, are associated with other simple nodes. Since a CVT node can never be the final answer node, every time a CVT node is encountered we need to extract the nodes associated with it. The key entities associated with each question is specified making it easier to start the graph exploration process (entity linking is a separate challenging topic not addressed here).  

\subsection{MetaQA}
This dataset contains questions in the movie domain \citep{zhang2017variational} and answer entities are up to 3 hops away from the topic entities on a movie KG. We focus only on the 3-hop examples that have the highest complexity in the MetaQA dataset. The graph is defined as a triplet (total 134,741) where each node is one of the following seven categories: movie, director, actor, writer, genre, language or release year. Relations to other nodes are defined as one of the following edge types: 'directed\_by', 'has\_genre', 'has\_imdb\_rating', 'has\_imdb\_votes', 'has\_tags', 'in\_language', 'release\_year', 'starred\_actors' and 'written\_by'. First we process the triplets to create movie specific dictionary where all relations for a specific movie is aggregated. Thus, for the movie 'Kismet' the dictionary has the following information:
\begin{verbatim}
{'directed_by': ['William Dieterle'],
 'written_by': ['Edward Knoblock'],
 'starred_actors': ['Marlene Dietrich',
  'Edward Arnold',
  'Ronald Colman',
  'James Craig'],
 'release_year': ['1944'],
 'in_language': ['English'],
 'has_tags': ['bd-r']}
\end{verbatim}

There are 114,196 train and 14,274 test examples. For each example there is 'key' that specifies the pattern of the query path required to be followed to arrive at the answer node. There are 15 such path types, e,g., 
movie\_to\_actor\_to\_movie\_to\_year, movie\_to\_writer\_to\_movie\_to\_director, etc. However, the same set of paths/patterns are present in both the train and test dataset. Thus, it is relatively simple to predict the path for a test example using few train examples (few-shot) followed by traversing the KG along this path to get the final result. We apply LLM to identify the right query path with and without few-shot examples taken from the training dataset. The few-shot examples are dynamically chosen for each question based on the embedding vector of the current question and all the pre-computed embedding vectors of the training set questions. Another simpler approach is to consider only the few-shot examples retrieved based on cosine similarity and suggest the most frequent query path found in these examples. Table~\ref{tab:metaqa_3hop} shows the results from both zero-shot and few-shot LLM applications and the majority query path based solution. 

\subsection{ComplexWebQuestions}
This dataset is created by Talmor and Berant \citep{talmor-berant-2018-web} that contains questions in the English language and is an extended version of the WebQuestions with 27,639 train, 3,519 dev and 3,531 test examples. This dataset requires complex operations, such as up to 4 hops and satisfying temporal constraints. The dataset has 2,429,346 Freebase entities and 6,649 relations (considerably larger than the WebQuestions dataset). Similar to the WebQuestions, we have created a vector database of 27k training examples to dynamically retrieve top-5 examples for each test example. However, this process does not pay attention to the complexity of the question (e.g., number of required hops) but retrieves based on semantic similarity and entities present in the question. Further, different examples have different numbers of key entities although all of them are not relevant (for example, common nouns like country). Another characteristic of this dataset is that the questions are not paraphrased properly, e.g., “Lou Seal is the mascot for the team that last won the World Series when?”. 
\begin{table*}
\centering
\begin{tabular}{lc}
\hline
Approach (WebQSP) & Hits@1 \\
\hline
EmbedKGQA,~\cite{saxena-etal-2020-improving} & 72.5 \% \\
StructGPT,~\cite{jiang-etal-2023-structgpt} (Few-shot, 15) & 72.6 \% \\
ReTraCk,~\cite{chen-etal-2021-retrack} & 74.6 \% \\
UniKGQA,~\cite{jiang2023unikgqa} & 77.2 \% \\
\hline
Ours \\
ChatGPT (Zero-shot) & 72.40\% \\
ChatGPT (Few-Shot, 5) & {\bf 80.20\%} \\ \hdashline
GPT-4 (Zero-shot) & 83.42 \% \\
GPT-4 (Few-shot) & 85.32 \% \\
\hline
\end{tabular}
\caption{Results from different approaches of using LLM for WebQSP question answering.}
\label{tab:webqsp}
\end{table*}

\begin{table*}
\centering
\begin{tabular}{lc}
\hline
Approach (MetaQA-3hop) & Hits@1 \\
\hline
UniKGQA,~\cite{jiang2023unikgqa} & 99.9 \% \\
StructGPT,~\cite{jiang-etal-2023-structgpt}  (Few-shot, 32) & 87.0 \% \\
\hline
Ours \\
ChatGPT (Zero-shot) & 82.74\% \\
ChatGPT (Few-Shot, 5) & {\bf 98.68 \%}  \\
\hline
\end{tabular}
\caption{Results from different approaches of using LLM and supervised training for MetaQA (3-hop) question answering.}
\label{tab:metaqa_3hop}
\end{table*}

\begin{table*}
\centering
\begin{tabular}{lc}
\hline
Approach (CWQ) & Hits@1 \\
\hline
Seq2seq + DKG, ~\cite{sen-etal-2023-knowledge} & 40.38 \% \\
SR+NSM+E2E,~\cite{He_2021} & 49.30 \% \\
UniKGQA,~\cite{jiang2023unikgqa} & 51.20 \% \\
\hline
Ours \\
ChatGPT (Zero-shot) & 20.50 \% \\
ChatGPT (Few-Shot, 5) & {\bf 52.59 \%}  \\
\hline
\end{tabular}
\caption{Results from different approaches of using LLM and supervised training for the ComplexWebQSP dataset.}
\label{tab:cwq}
\end{table*}

\begin{table*}
\centering
\begin{tabular}{lcc}
\hline
Approach (LC-QuAD 1.0) & Hits@1 & Macro-F1 \\
\hline
Core chain candidate Ranking, \cite{maheshwari2018learning} \\
CNN Encoder & & 50.0 \% \\
BiLSTM Encoder & & 55.0 \% \\
Slot-Matching Encoder  & & {\bf 71.0} \% \\
UNIQORN, \cite{pramanik2023uniqorn} & 26.5 \% & \\
\hline 
Ours \\
ChatGPT (Zero-shot) & 28.0 \% \\
ChatGPT (Few-shot, 5) & {\bf 61.8 \%} & 59.5 \%  \\
\hline
\end{tabular}
\caption{Results from different approaches of using LLM and supervised training for the LC-QuAD-1.0 dataset.}
\label{tab:lcquad}
\end{table*}

\subsection{LC-QuAD V1 and V2}
Large-Scale Complex Question Answering Dataset (LC-QuAD) v1.0 \citep{trivedi2017lc} is a KGQA dataset with 5000 example pairs of question and the corresponding SPARQL query. The dataset is based on the 2016 DBpedia graph and can be queried through public endpoints \footnote{https://dbpedia.org/sparql} or locally created endpoints using downloaded DBpedia files. We have created a local endpoint using Virtuoso cloud server with the 04-2016 DBpedia files for run-time query. 

LC-QuAD v2 \citep{lcquad_2} is a continuation of the previous dataset with more examples (30,000 question-query pairs) and compatible with both Wikidata and DBpedia 2018 KG. We use the Wikidata public endpoint\footnote{https://query.wikidata.org/sparql} to evaluate our predicted SPARQL queries. In addition, a list of 22,792 entity QIDs and 3627 predicates (PIDs) are provided without any description. We query Wikipedia\footnote{https://www.wikidata.org/wiki/Special:EntityData/} to get the corresponding entity and predicate descriptions and store them separately. We use these descriptions as documents for Retrieval Augmented Generation (RAG) for predicting the relevant nodes and edges, as defined in Fig.~\ref{fig:sp}. We also use the training split for creating a vector database to be used for few-shot examples with test set questions. Unlike the previous datasets, here we do not know the key entities, $\mathcal{T}_q$. There are three questions given for each example as shown below: 
\begin{itemize}
    \item “question”: "What periodical literature does Delta Air Lines use as a moutpiece?"
    \item "NNQT\_question": "What is the {periodical literature} for {mouthpiece} of {Delta Air Lines}",
    \item "paraphrased\_question": "What is Delta Air Line's periodical literature mouthpiece?"
\end{itemize}
and we use only the string associated with the “question” key and never the annotated one. 

\begin{table*}
\centering
\begin{tabular}{lc}
\hline
Approach (LC-QuAD 2.0) & Hits@1 \\
\hline
Flan-T5-XL, \cite{sen-etal-2023-knowledge} & 22.82 \% \\
UNIQORN, \cite{pramanik2023uniqorn} & 32.40 \% \\
\cite{dai2024counterintuitive} (only 2000 examples) \\
 ChatGPT - without triples & 16.42 \% \\
 ChatGPT - with triples & 51.71 \% \\
\hline 
Ours \\
ChatGPT (Entities Unknown) & 17.89 \% \\
ChatGPT (Entities known) & {\bf 56.98 \%}  \\
\hline
\end{tabular}
\caption{Results from different approaches of using LLM and supervised training for the LC-QuAD-2.0 dataset.}
\label{tab:lcquad2}
\end{table*}

\subsection{KQAPro}
KQAPro is the largest KGQA dataset with 94,376 train and 11,797 validation and test examples~\citep{cao-etal-2022-kqa}. It is based on a dense subset of Wikidata \citep{wikidata_2014} that includes multiple types of knowledge involving multi-hop inference, logical union and intersection, etc. Each example in the dataset provides a SPARQL query and a KoPL logical form along with a set of candidate answers. However, unlike LC-QuAD, the entities (and predicates) in the SPARQL queries are not coded by Q-ids (and P-ids) but their original form is retained, e.g., "Georgia national football team" or <pred:unit>. There is no separate list of entities and predicates unlike LC-QuAD, however, executions of SPARQL queries generate entities in terms of Q-ids and other literal values. We parse all the SPARQL queries to extract the relevant entities and predicates and use their descriptions as documents for RAG. Similar to LC-QuAD, we also use the training split for creating a vector database to be used to generate few-shot examples while evaluating the test set. In addition, the key entities, $\mathcal{T}_q$ are not mentioned for any of the training/validation/test examples. Similar to LC-QuAD v1.0, we have created a local endpoint using Virtuoso cloud server to evaluate the predicted and golden SPARQL queries.

\begin{table*}
\centering
\begin{tabular}{lcc}
\hline
Approach (KQAPro) & Hits@1 & EM-Accuracy \\
\hline
BART KoPL, \cite{cao-etal-2022-kqa} & & 90.55 \% \\
GraphQ-IR, \cite{nie-etal-2022-graphq} & & {\bf 91.70} \% \\
\hline
\cite{dai2024counterintuitive} (only 2000 examples) \\
 ChatGPT - without triples & 15.77 \% \\
 ChatGPT - with triples & 54.19 \% \\
 ChatGPT, \cite{tan2023chatgpt} & & 47.93 \% \\
\hline 
Ours \\
ChatGPT (Entities Unknown) & 71.85 \% & 77.92 \% \\
ChatGPT (Entities known) & {\bf 72.90 \%} & 72.75 \%  \\
\hline
\end{tabular}
\caption{Results from different approaches of using LLM and supervised training for the KQAPro dataset.}
\label{tab:kqapro}
\end{table*}

\section{Results}
For all datasets we report Hit@1 metric (as is done in other similar studies), i.e., if the LLM generated answer contains at least one correct answer entity we consider that question is answered correctly. For webQSP we have also converted both the ground truth and predicted answers into lower cases before comparing them. This is not required, however, for LC-QuAD (V1 and 2) and KQAPro, since the answers are retrieved by executing SPARQL queries. 

Table~\ref{tab:webqsp} shows the performance of our IR-LLM approach as applied to the WebQSP dataset. The first four rows are some of the existing baselines where UniKGQA~\citep{jiang2023unikgqa} shows the best performance of 77.2\%. The rest of the four rows show results from our experiments where with few-shot learning we achieves 80.20\% and 85.32\% for GPT-3.5-Turb and GPT-4, respectively. Thus, IR-LLM based in-context learning surpassed supervised learning based performance. It is to be noted that StructGPT~\citep{jiang-etal-2023-structgpt} with few-shot (15 examples) learning achieves only 72.6\%. 

Table~\ref{tab:metaqa_3hop} shows the results for the Meta-QA dataset. As can be seen, our approach results in significantly higher Hit@1 than that of StructGPT and close to the UniKGQA performance. It is to be noted that we have only added 5 few-shot examples, whereas, for StructGPT it was 32. Further, even though the number of hop is 3 IR-LLM achieves better result than that of WebQSP since the complexity is less for this dataset.
% To understand the capability of LLM to generate the relevant query path we randomly select 10 paths (out of 15) and make them available for creating few-shot examples while the rest 5 paths are reserved for testing. We randomly sample 100 examples from the training split for each path and generate a dataset of 1000 examples. We evaluate only those examples in the test split that belong to one of the 5 paths. There are 4335 such examples in the test set and GPT-3.5 correctly identifies the right query path for 3941 cases resulting in an accuracy of 90.91\%.   

For the ComplexWebQSP~\citep{talmor-berant-2018-web} dataset, the performance is shown in Tab.~\ref{tab:cwq}. All the comparisons are with other supervised learning based approaches as we could not find any other in-context learning based work on this dataset. This is the most complex dataset where even the supervised training achieves no more than 51.2\%. Our IR-LLM approach, however, surpasses that and sets a new SOTA performance at 52.59\%. The impact of dynamic few-shot is evident here as it improves the accuracy from 20.50\%.  

On the SP-LLM approach, we evaluate LC-QuAD (both V1.0 and 2.0) and KQAPro datasets and compare with other results as available in the literature. For all the SP-LLM based results, we have two scenarios, (a) assuming that the relevant entities and attributes are not known and should be inferred from the query and (b) assuming that the relevant entities are known a priori. Similar study is carried out by \citet{dai2024counterintuitive} where they assume that either the triples are not known (similar to our first scenario) or known by extracting these entities from SPARQL queries (our second scenario). However, they have only reported results for LC-QuAD v2.0. The results for LC-QuAD v1.0 is reported in Tab.~\ref{tab:lcquad} where our approach results in the best Hits@1 compared to the other methods. However, if we also compare the F1-score we can see that the slot-matching model (fine-tuned) of \citet{maheshwari2018learning} has significantly better performance although other encoders (CNN and BiLSTM) may result in substantially lower F1-score. 

For LC-QuAD V2.0~\citep{lcquad_2} the results are shown in Tab.~\ref{tab:lcquad2} where we report only the Hits@1 metric. As can be seen for both the scenarios of known and unknown entities, our approach performs better than that of \citet{dai2024counterintuitive} (17.89\% vs. 16.42\% and 56.98\% vs. 51.71\%). There are few other studies based on fine-tuned models that lag significantly behind our ChatGPT based approach. It can also be seen that the knowledge of the key entities is critical to achieve a reasonable performance. 

The performance on the KQAPro~\citep{cao-etal-2022-kqa} dataset is shown in Tab~\ref{tab:kqapro}. To compare with the other available results, we also report the exact match (EM) accuracy, in addition to Hits@1. While our SP-LLM approach results in 72.9\% Hits@1 which is better than the other comparable approach of \citet{dai2024counterintuitive}, in terms of EM accuracy, our performance (76.96\%) is better than the previous ChatGPT attempt (47.93\%) \citep{tan2023chatgpt} but significantly behind the other fine-tuned model based results, namely, BART KoPL~\citep{cao-etal-2022-kqa} and GraphQ-IR~\citep{nie-etal-2022-graphq}. We can also see that there is very little difference in our results between the known and unknown entity scenarios (72.90 vs. 71.98\%), presumably, the few-shot examples are sufficient for ChatGPT to construct the SPARQL query. This brings us to the nature of the distributions of KQAPro where 117,970 examples have 24,724 unique answers. In addition, only "30\% of the test set are not seen in training" \citep{cao-etal-2022-kqa} indicating significant overlap between the train and test split. This can explain why the few-shot examples play a critical role in our results and how the fine-tuned models achieved very high EM-accuracy.

% \begin{table*}
% \centering
% \begin{tabular}{lccccc}
% \hline
% \multicolumn{1}{c}{Title} & \multicolumn{5}{c}{Best Results} \\ 
% \hline
%  & MetaQA (3hop) & WebQSP & ComplexWebQSP & LC-QuAD & MINTAKA \\
% \hline
% \cite{guo2024knowledgenavigator} & 95\% & 83.5\% &  \\
% \cite{sen-etal-2023-knowledge} & - & 59.79\% & 40.38\% & 22.82\% & 37.90 \\
% UniKGQA, \cite{jiang2023unikgqa} & 99.9\% & 77.2\% & 51.2\% \\
% StructGPT, \cite{jiang-etal-2023-structgpt} & 87.0\% & 72.6\% \\
% RRA, \cite{wu2023retrieverewriteanswer} & - & 79.36\% \\
% ZSKGQA, \cite{baek2023knowledgeaugmented} & - & 73.89\% & & & 56.86\% \\
% \hline
% \end{tabular}
% \caption{Summary of previously obtained results for different KGQA datasets and approaches.}
% \end{table*}

\section{Conclusion}
In this study, we show how in-context learning can be utilized for both IR and SP based approaches for KG question answering. We introduce a novel pipeline that iterates over LLM extracted information (nodes or edges) and decides whether to continue the search or not. We show that this approach (IR-LLM) performs better than the existing solution and shows 8\% and 11\% improvement over the WebQSP and MetaQA-3hop datasets, respectively. On the CWQ dataset, we achieve a new SOTA performance surpassing the existing supervised training based results. For the SP-LLM approach based results, we get SOTA performance on LC-QuAD V1.0, 2.0 and KQAPro on the Hits@1 metric, however, for LC-QuaD V1.0 and KQAPro we lag behind other fine-tuned models on other metrics (EM-accuracy and F1-score). The future line of work can be directed to achieve better performance on these additional metrics as well. 

% \section*{Acknowledgments}
% This was was supported in part by......

%Bibliography
% \bibliographystyle{unsrt}
\bibliographystyle{acl_natbib}
\bibliography{references}

\end{document}